\documentclass{article}
\usepackage{tikz}
\usepackage{pgfplots}
\pgfplotsset{compat=1.17}
\usetikzlibrary{shapes.geometric, arrows, positioning}
\usetikzlibrary{decorations.pathreplacing, shapes.geometric, arrows, positioning}
\definecolor{maroon}{rgb}{0.5, 0, 0}
\definecolor{darkblue}{rgb}{0.0, 0.0, 0.55}
\definecolor{pblue}{HTML}{2980FF}
\definecolor{plightred}{HTML}{FF7676}
\definecolor{pdarkred}{HTML}{FF000D}
\usepackage{amsmath}
\usepackage{fontawesome}
\tikzstyle{process} = [rectangle, rounded corners, minimum width=2.5cm, minimum height=1cm, text centered, draw=black, fill=blue!20]
\tikzstyle{process1} = [rectangle, rounded corners, minimum width=0.8cm, minimum height=0.8cm, text centered, draw=black, fill=darkblue!50]
\tikzstyle{process2} = [rectangle, rounded corners, minimum width=0.8cm, minimum height=0.8cm, text centered, draw=black, fill=maroon!50]
\tikzstyle{arrow} = [thick,->,>=stealth]
\usepackage{url}
\usepackage{booktabs}
\usepackage{subcaption}
\usepackage{multirow}

\usepackage{arxiv}

\usepackage[utf8]{inputenc} % allow utf-8 input
\usepackage[T1]{fontenc}    % use 8-bit T1 fonts
\usepackage{hyperref}       % hyperlinks
\usepackage{url}            % simple URL typesetting
\usepackage{booktabs}       % professional-quality tables
\usepackage{amsfonts}       % blackboard math symbols
\usepackage{nicefrac}       % compact symbols for 1/2, etc.
\usepackage{microtype}      % microtypography
\usepackage{lipsum}		% Can be removed after putting your text content
\usepackage{graphicx}
\usepackage[numbers]{natbib}
\usepackage{doi}

\title{Accelerated Gradient-based Design Optimization Via Differentiable Physics-Informed Neural Operator: \newline A Composites Autoclave Processing Case Study}

\author{{\hspace{1mm}Janak M. Patel}\thanks{Corresponding author.} \\
        Applied Research, Quantiphi\\
        Marlborough, MA 01752, USA\\
	\texttt{janak.patel@quantiphi.com} \\
	\And
	{\hspace{1mm}Milad Ramezankhani}\\
	Applied Research, Quantiphi\\
        Marlborough, MA 01752, USA\\
	\texttt{milad.ramezankhani@quantiphi.com} \\
        \And
	{\hspace{1mm}Anirudh Deodhar} \\
	Applied Research, Quantiphi\\
        Marlborough, MA 01752, USA\\
	\texttt{anirudh.deodhar@quantiphi.com} \\
        \And
	{\hspace{1mm}Dagnachew Birru} \\
	Applied Research, Quantiphi\\
        Marlborough, MA 01752, USA\\
	\texttt{dagnachew.birru@quantiphi.com} \\
}

\renewcommand{\shorttitle}{Accelerated Gradient-based Design Optimization Via Physics Informed Neural Operator}

\hypersetup{
pdftitle={A template for the arxiv style},
pdfsubject={q-bio.NC, q-bio.QM},
pdfauthor={David S.~Hippocampus, Elias D.~Striatum},
pdfkeywords={First keyword, Second keyword, More},
}

\begin{document}
\maketitle

\begin{abstract}
Simulation and optimization are crucial for advancing the engineering design of complex systems and processes. Traditional optimization methods require substantial computational time and effort due to their reliance on resource-intensive simulations, such as finite element analysis, and the complexity of rigorous optimization algorithms. Data-agnostic AI-based surrogate models, such as Physics-Informed Neural Operators (PINOs), offer a promising alternative to these conventional simulations, providing drastically reduced inference time, unparalleled data efficiency, and zero-shot super-resolution capability. However, the predictive accuracy of these models is often constrained to small, low-dimensional design spaces or systems with relatively simple dynamics. To address this, we introduce a novel Physics-Informed DeepONet (PIDON) architecture, which extends the capabilities of conventional neural operators to effectively model the nonlinear behavior of complex engineering systems across high-dimensional design spaces and a wide range of dynamic design configurations. This new architecture outperforms existing state-of-the-art models, enabling better predictions across broader design spaces. Leveraging PIDON's differentiability, we integrate a gradient-based optimization approach using the Adam optimizer to efficiently determine optimal design variables. This forms an end-to-end gradient-based optimization framework that accelerates the design process while enhancing scalability and efficiency. We demonstrate the effectiveness of this framework in the optimization of aerospace-grade composites curing processes achieving a \(3 \times\) speedup in obtaining optimal design variables compared to gradient-free methods. Beyond composites processing, the proposed model has the potential to be used as a scalable and efficient optimization tool for broader applications in advanced engineering and digital twin systems.
\end{abstract}

% keywords can be removed
\section{Introduction}
Simulation and optimization play a pivotal role in modern engineering, serving as essential tools for analyzing, designing, and enhancing complex systems and processes. Engineers leverage simulation models to explore design spaces, evaluate performance, and guide decision-making through optimization. Traditionally, numerical simulation methods, such as finite element analysis, have been widely used in combination with optimization algorithms to determine optimal design variables \cite{struzziero2017multi,dolkun2018optimization,tang2022multi}. However, these methods are computationally expensive and time-intensive, making them less practical for real-time or iterative-based design processes \cite{ZIMMERLING2022110423}. To address these challenges, machine learning-based surrogate models have been developed as alternatives to reduce computational costs \cite{pfrommer2018optimisation, tifkitsis2018stochastic,yuan2021multi}. Once trained, these models provide fast predictions, enabling near real-time optimization. However, surrogate models often require large datasets for training, which is prohibitive in many industrial scenarios with data scarcity, resulting in unreliable or physically implausible predictions.

Recently, Physics-Informed Neural Networks (PINNs) are introduced \cite{raissi2019physics}, which incorporate governing partial differential equations (PDEs) into the loss function of a neural network. This allows the model to learn directly from the underlying physics, combining the strengths of numerical methods with machine learning-based surrogate modeling. While PINNs offer a promising alternative for solving PDEs without requiring labeled data, they are typically trained on fixed initial and boundary conditions (ICs/BCs) \cite{wang2022mosaic}. This limitation makes them less practical for iterative design optimization tasks, where system configurations (e.g., different BCs) frequently change. In contrast, neural operators such as DeepONet \cite{lu2019deeponet} and Fourier neural operator (FNO) \cite{li2020fourier}, which directly map infinite-dimensional input-output function spaces on bounded domains, are capable of learning a family of PDEs rather than a single instance \cite{boulle2023mathematical,kovachki2023neural}. Inspired by PINNs, Physics-Informed Neural Operators (PINOs) extend this capability by incorporating physics constraints into the loss function, allowing them to solve PDE families without relying on data \cite{wang2021learning,goswami2023physics}. Unlike PINNs, PINOs enable real-time predictions under varying conditions, making them ideal for design optimization. Despite the advantages of PINOs in capturing system dynamics under varying conditions, existing architectures may struggle with high-dimensional and large design spaces for real-world engineering problems. Additionally, traditional gradient-free optimization methods, such as Particle Swarm Optimization (PSO) \cite{pyswarmsJOSS2018} and Genetic Algorithms (GA) \cite{solgi_geneticalgorithm}, often require a large number of function evaluations, making them computationally expensive and inefficient for high-dimensional optimization \cite{allen2022physical}. 

To address these challenges, we propose an accelerated gradient-based optimization framework tailored for advanced engineering design. Specifically, Our work focuses on composite materials processing, where achieving an optimal cure cycle is critical for ensuring material quality. Composite materials are extensively used in aerospace, automotive, and marine industries due to their high strength, durability, and lightweight properties \cite{li2020cure}. These materials undergo a polymerization process governed by a temperature and pressure cycle known as the cure cycle \cite{strong2008fundamentals}. The cure cycle must be optimized to ensure uniform resin curing while minimizing residual stress and deformation \cite{hubert2001cure}. Since changes in part geometry or material properties often require adjustments to the cure cycle, optimization becomes essential for ensuring high-quality and reliable composites.

Numerous studies have utilized PINO models to simulate the thermochemical evolution of composites during autoclave curing. For instance, a Physics-guided Neural Operator is developed for composite manufacturing \cite{chen2023physics}. Similarly, physics-informed FNO was applied to analyze the curing process of carbon-fiber composites \cite{meng2023novel}. Furthermore, a Physics-Informed DeepONet, enhanced with advanced features such as nonlinear decoders and curriculum learning is developed for modeling complex thermochemical processes \cite{ramezankhani2025advanced}. However, the predictive performance of these models remains satisfactory only when applied to single-variable designs, limited design spaces, or systems with low behavioral complexity. This limitation may arise from the inherent constraints of the original neural operator architecture in handling nonlinear and stiff problems, as well as complex physics-informed loss landscape \cite{krishnapriyan2021characterizing} and neural network's spectral bias \cite{rahaman2019spectral}. To address these challenges and ensure that the developed surrogate model remains accurate when exposed to high-dimensional and large design spaces for optimization tasks, we propose an advanced Physics Informed DeepONet (PIDON) architecture. In particular, it incorporates domain decomposition with separate DeepONets allocated to each temporal subdomain, as well as input coordinate normalization to mitigate spectral bias \cite{moseley2023finite}. Furthermore, each subdomain is equipped with a nonlinear decoder to better capture the complex dynamics of the PDE. The proposed PIDON model serves as an efficient and accurate surrogate for inverse design, enabling near real-time spatiotemporal predictions for given design conditions. This capability enhances design space exploration and accelerates the identification of optimal design variables to achieve desired material properties. Leveraging the differentiability of PIDON \cite{li2024physics}, we utilize a gradient-based optimization technique, which significantly reduces the number of function evaluations and improves scalability for larger design spaces compared to gradient-free methods \cite{allen2022physical}. The Adam optimizer \cite{kingma2014adam} is used to obtain the optimal design variables. We benchmark this framework against gradient-free methods, including PSO and GA. The main contributions of this paper are summarized as:
% \begin{itemize}  
%     \item \textbf{AI-driven Accelerated Design Optimization framework:} Development of an end-to-end framework for accelerated AI-driven design optimization of the curing process for composite materials
%     \item \textbf{Novel PIDON Architecture:} Development of PIDON model with domain decomposition, separate input normalization for each subdomain and a nonlinear decoder
%     \item \textbf{Gradient-Based Optimization:} Integration of the Adam optimizer with the differentiable PIDON model, achieving efficient and scalable optimization compared to traditional gradient-free methods
% \end{itemize}  
\begin{itemize}  
    \item Design and implement an improved PIDON model for composites manufacturing, enabling an accurate and accelerated exploration of high-dimensional and large design spaces.
    \item Develop an end-to-end AI-driven design optimization framework enabling \(3\times\) speedup over conventional gradient-free approaches.
\end{itemize}  
\section{Methodology}
\subsection{Composites autoclave processing}
\label{composites}
\begin{figure*}[t]  
    \centering
    \includegraphics[width=1\textwidth]{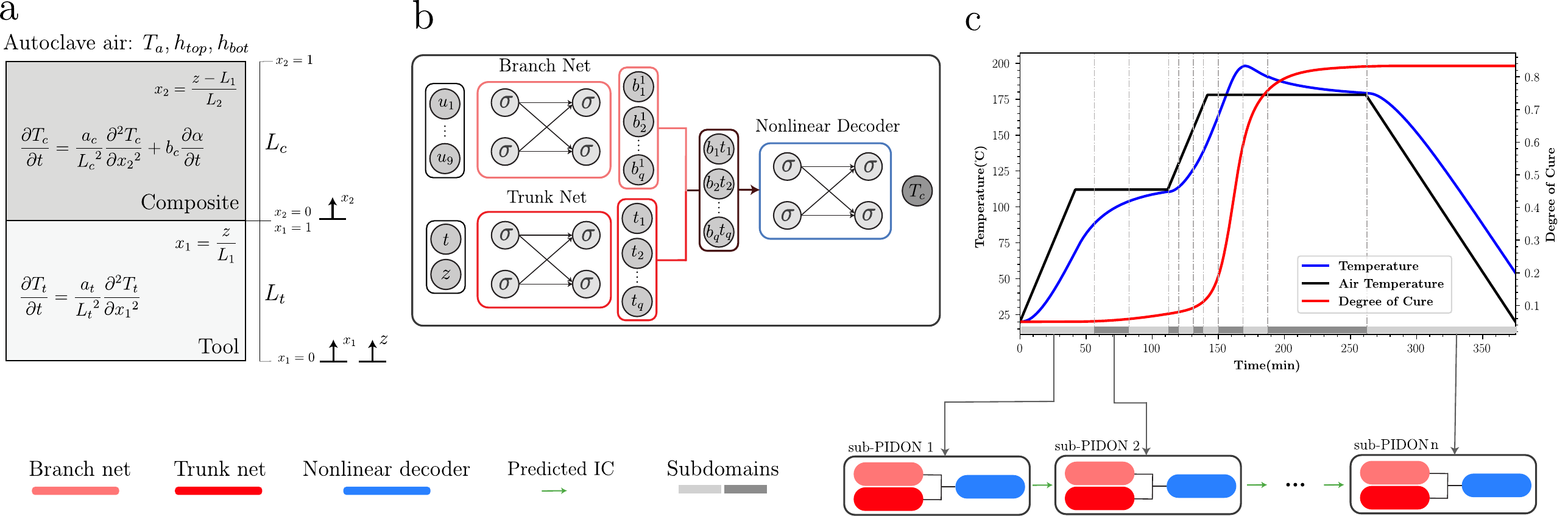} 
    \caption{a) Schematic representation of the composite-tool system inside an autoclave, including local coordinates $x_1$ and $x_2$ (adopted from \cite{ramezankhani2025advanced}). b) Architecture of the proposed sub-PIDON model for predicting part temperature \(G^{T_c}\). The same architecture is utilized for other output variables, including DOC \(G^{\alpha}\) and tool temperature \(G^{T_t}\). c) Illustration of the proposed PIDON framework with a nonlinear decoder and domain decomposition, designed for thermochemical analysis during the composite curing process.}     % Caption for the figure
    \label{fig:system}                             % Label for referencing the figure
\end{figure*}

The autoclave processing in the manufacture of composites involves placing resin-impregnated fibers and a tool into an autoclave, where the system undergoes a cure cycle to achieve the desired material properties, as illustrated in Figure \ref{fig:system}.a. The temperature distribution within the part ($T_c$) and tooling ($T_t$), along with the progression of the resin's degree of cure (DOC), are critical state variables in composites systems. The thermochemical behavior of a 1D composite-tool system in an autoclave is described by the following one-dimensional anisotropic heat conduction equation:
\begin{equation}
\begin{aligned}
\frac{\partial T_t}{\partial t} &= a_t \frac{\partial^2 T_t}{\partial z^2}, \quad z \in [0, L_t] \\
\frac{\partial T_c}{\partial t} &= a_c \frac{\partial^2 T_c}{\partial z^2} + b_c \frac{\partial \alpha}{\partial t}, \quad z \in [L_t, L_c]
\end{aligned}
\end{equation}
Here, $T$ is the temperature, $\alpha$ is DOC, $L_c$ is the material length, $L_t$ is the tool length, and $t$ and $z$ are the spatiotemporal coordinates. The subscripts $t$, $c$, and $r$ represent the tool, composite part, and resin, respectively. The parameter $a$ denotes the thermal diffusivity, and $b$ represents the heat generation coefficient. For the curing process of a composite system with thermoset resin, the cure rate is governed by the resin's cure kinetics, which are typically represented by an ordinary differential equation. For the AS4/8552 epoxy resin system, it can be expressed as follows\citep{hubert2001cure}:
\begin{equation}
\frac{\partial \alpha}{\partial t} = A \exp\left(-\frac{\Delta E}{RT}\right) \frac{1}{1 + \exp\left(C(\alpha - (C_0 + C_T T))\right)} \alpha^m (1 - \alpha)^n
\end{equation}
where, $\Delta E$ represents the activation energy, $R$ is the gas constant, and $C_0$, $C_T$, $m$, $n$, and $A$ are experimentally determined constants and parameter. The values used for this study can be found in \citep{johnston1997integrated}.
Considering the convective heat transfer between the autoclave air $T_a$ and the composite system, the boundary conditions are governed by:
\begin{equation}
\begin{aligned}
(T_a - T_c \big|_{z=L_c}) &= \frac{k_c}{h_{\text{top}}} \frac{\partial T_c}{\partial z} \Bigg|_{z=L_c} \\
(T_t \big|_{z=0} - T_a) &= \frac{k_t}{h_{\text{bot}}} \frac{\partial T_t}{\partial z} \Bigg|_{z=0}
\end{aligned}
\end{equation}
Here, $h_{\text{top}}$ and $h_{\text{bot}}$ are the convective heat transfer coefficients (HTCs) on the top and bottom surfaces of the composite-tool system, respectively, while $k_c$ and $k_t$ represent the thermal conductivity of the composite and tool, respectively. The initial temperature of the part is typically assumed to be uniform throughout. For this study, we assume an initial temperature of 20°C. The initial DOC is assumed to be either zero or a very small value for an uncured part and in this study, is set to 0.05. The part thickness is fixed for this study, while the tool thickness is included as one of the design variables. To manage inconsistencies in the total system length and interface location resulting from variations in tool thickness, local coordinates are introduced, following the approach outlined by \citep{ramezankhani2025advanced}.

A typical two-hold cure cycle consists of two distinct stages where the air temperature is held constant at specific levels for predetermined durations. This approach enables a controlled progression of the curing process, allowing optimal material properties to be achieved while minimizing potential issues such as under-curing or over-curing \citep{fabris2018framework}. The cure cycle is defined using six key design variables: heating rates \(r_1\) and \(r_2\) (the rate at which the air temperature inside the autoclave increases), hold durations \(hd_1\) and \(hd_2\) (the time periods for which the temperature is maintained constant at the respective hold temperatures) and hold temperatures \(ht_1\) and \(ht_2\) (the target temperatures held during the first and second stages).
\subsection{Differentiable Simulator: Physics Informed DeepONet} 
The architecture of DeepONet has two primary components: a branch and a trunk network \citep{lu2019deeponet}. The branch network takes as input the sensor point evaluations $u$ $=$ $[u(x_1), u(x_2)$,$\dots, u(x_m)]$  and generates a feature representation $b$ $=$ $[b_1, b_2, \dots, b_q]^T \in \mathbb{R}^q$ as its output. Similarly, the trunk network encodes the spatiotemporal coordinates of the PDE system \( y \) into a feature embedding \( t = [t_1, t_2, \dots, t_q]^T \in \mathbb{R}^q \), with the same dimensionality as the output of the branch network. The outputs from these networks are combined using an element-wise product operation, followed by summation, to produce the final output of the DeepONet:
\begin{equation}
G_\theta(u)(y) = \sum_{k=1}^{q} b_k t_k + b_0
\end{equation}
Data-driven DeepONet often requires large amounts of training data, which can be difficult to access in many real-world engineering applications. The physics-informed variant of DeepONet, namely PIDON, integrates governing equations directly into the loss function as regularizers, eliminating DeepONet's reliance on data. Specifically, the output of the PIDON is constrained to satisfy the governing equations by minimizing the loss function:
\begin{equation}
\mathcal{L}(\theta) = \mathcal{L}_{IC}(\theta) + \mathcal{L}_{BC}(\theta) + \mathcal{L}_{physics}(\theta)
\end{equation}
where $\mathcal{L}_{IC}$, $\mathcal{L}_{BC}$, and $\mathcal{L}_{physics}$ represent the initial condition (IC) loss, boundary condition (BC) loss, and physics loss, respectively. Assuming a constant initial condition and a Robin boundary condition in the introduced composites case study, \(\mathcal{L}_{IC}\) and \(\mathcal{L}_{BC}\) can be expressed as:
\begin{equation}
\mathcal{L}_{IC}(\theta) = \frac{1}{NQ_{ic}} \sum_{i=1}^{N} \sum_{j=1}^{Q_{ic}} \left| G_{\theta}(u^{(i)})(y^{(i)}_j) - s^{(i)}(y^{(i)}_j) \right|^2
\end{equation}
\begin{equation}
\mathcal{L}_{BC}(\theta) = \frac{1}{NQ_{bc}} \sum_{i=1}^{N} \sum_{j=1}^{Q_{bc}} \left| \alpha G_{\theta}(u^{(i)})(y^{(i)}_j) + \beta \nabla G_{\theta}(u^{(i)})(y^{(i)}_j) - \gamma \right|^2
\end{equation}

Here, \(u^{(i)}\) denotes the \(i\)-th input function, \(y_j^{(i)}\) represents the \(j\)-th collocation point , and \(G_\theta\) is the output of the PIDON. The term \(s^{(i)}(y_j^{(i)})\) corresponds to the solution of the partial differential equation (PDE) at \(y_j^{(i)}\), conditioned on the \(i\)-th input function. For the Robin boundary condition, \(\alpha\), \(\beta\), and \(\gamma\) are non-zero constants determined by the physical characteristics of the problem. The physics loss \(\mathcal{L}_{physics}\) is defined as:

\begin{equation}
\mathcal{L}_{physics}(\theta) = \frac{1}{NQ} \sum_{i=1}^{N} \sum_{j=1}^{Q} \left| \mathcal{N}(u^{(i)}(x), G_{\theta}(u^{(i)})(y^{(i)}_j)) \right|^2
\end{equation}

Here, \( \mathcal{N} \) denotes the nonlinear differential operator. The parameter \( N \) represents the number of distinct input function combinations sampled from the design space, whereas \( Q \) indicates the number of residual points employed to enforce the physical constraints. These \( N \) and \( Q \) are hyperparameters that can be tuned to balance the model's performance and computational efficiency. DeepONet is considered differentiable because it is a neural network-based approach that learns a solution operator, and a neural network is a differentiable function \cite{lu2019deeponet}.

\subsubsection{Proposed Architecture of PIDON}
We introduce a novel PIDON architecture designed to model the highly nonlinear dynamics of composite-tool systems during the curing process. The key components of the framework are outlined below. 

\textbf{Network architecture}.  This architecture incorporates a branch network, a trunk network, and a nonlinear decoder, drawing inspiration from advancements in operator learning \citep{seidman2022nomad, haghighat2024deeponet, ramezankhani2025advanced}. As illustrated in Figure \ref{fig:system}.b, the branch network is designed with multi-input functionality, accommodating 9 input parameters: six cure cycle parameters (\(r_1, r_2, hd_1, hd_2, ht_1, ht_2\)), two equipment design parameters(\(h_{top}, h_{bot}\), representing the convective heat transfer coefficients on the top and bottom surfaces of the composite-tool system, respectively), and one tool design parameter (tool thickness, \(L_t\) ). The trunk network takes spatio-temporal coordinates \(t\) and \(z\) as inputs. Standard DeepONet architectures with linear decoders require a large output dimension for branch and trunk networks to model nonlinear systems, making them computationally expensive and ineffective in such scenarios. In this work, we incorporate a nonlinear decoder by introducing an additional neural network that takes as input the combined output of the branch and trunk networks and generates the final output of PIDON.

\textbf{Temporal domain decomposition}. The composites curing process is a time-consuming process which requires solving the corresponding PDE equations over a long temporal domain. This poses some training challenges due to spectral bias and training difficulties such as activation saturation from large input coordinates. A temporal domain decomposition strategy is proposed, where a single DeepONet is trained with initial conditions as additional input functions \cite{wang2023long}. This enables the model to independently learn solution operators for different subdomains while ensuring temporal continuity. However, inaccuracies in predicting initial conditions can introduce and propagate errors in the model predictions. Our framework improves temporal domain decomposition by employing multiple DeepONets with separate spatiotemporal input normalization for each subdomain. In particular, the temporal domain is divided into $n$ smaller temporal subdomains based on the system's physical characteristics, with each subdomain modeled by an independent DeepONet, termed \textbf{sub-PIDON} (Figure~\ref{fig:system}.c). This setup enables each sub-PIDON to accurately capture the dynamics specific to its respective subdomain while reducing errors associated with relying on a single model to represent both initial conditions and other input functions within the system. The incorporation of separate subdomain normalization alongside domain decomposition effectively mitigates spectral bias over extended temporal domains by ensuring that the solution frequency encountered within each subdomain remains low \citep{moseley2023finite}.

\textbf{Multi-output functionality}. The PIDON framework predicts three key state variables for the thermochemical evolution in the composites curing process: $T_c$,  $T_t$, and $\alpha$. Due to the system's complexity, a single DeepONet may not effectively learn all these variables. Instead, we develop three dedicated and decoupled neural operators: \(G^{T_c}\) for composite temperature, \(G^{T_t}\) for tool temperature, and \(G^{\alpha}\) for the DOC, with each DeepONet specifically designed for its respective variable while maintaining consistent architectural principles. Thus, for each subdomain (highlighted with light and dark grey in Figure~\ref{fig:system}.c), we train three sub-PIDON, each dedicated to an output variable.
\subsubsection{Training Procedure}
As depicted in Figure \ref{fig:system}.c, the proposed PIDON framework is trained sequentially across multiple temporal subdomains. The training process begins with the initialization and training of the first sub-PIDON module (\texttt{sub-PIDON 1}) using the system’s global IC across the design space (i.e., input functions). Upon completion of its training, this module generates predictions that serve as the local IC for the subsequent sub-PIDON module (\texttt{sub-PIDON 2}). In this manner, the IC for each subdomain is determined based on the predictions of the preceding sub-PIDON model, as indicated by the green arrows in Figure \ref{fig:system}.c. This iterative process continues until all sub-PIDON modules corresponding to the defined subdomains are fully trained. The subdomains are initially partitioned into segments of equal width; however, their widths are adaptively refined during training. Specifically, if the total training loss for a given subdomain fails to reach a predefined threshold, the subdomain is further subdivided into two smaller intervals, thereby simplifying the learning task for the corresponding sub-PIDON modules. This adaptive approach enables more substantial progression in regions of the time domain where nonlinearity is less pronounced, while ensuring that highly complex regions are modeled with finer resolution. As a result, the framework maintains both computational efficiency and predictive accuracy across varying levels of system complexity.
\subsection{AI-Driven Accelerated Design Optimization Framework}
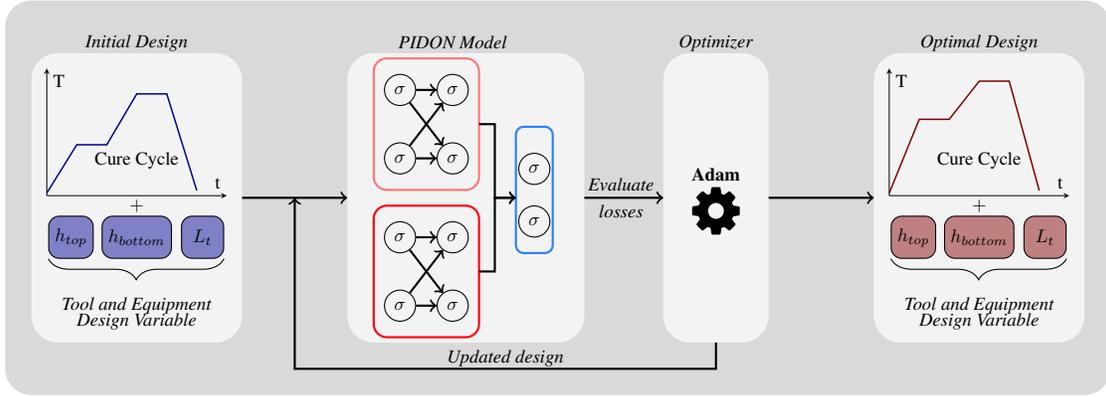
\begin{figure}[h!]
\centering
\begin{tikzpicture}[scale=0.7, transform shape]
% Create big box
\fill[gray, opacity=0.3,rounded corners=10pt] (-7.5,-5) rectangle (13.5,2.5);

% intial design box
\fill[white, opacity=0.7, rounded corners=10pt] (-7,-4) rectangle (-3,1.5);
% --- Line Plot ---
\node (plot) at (-5,0) {
\begin{tikzpicture}
    \begin{axis}[
        width=5cm, height=4cm,
        xlabel={t}, ylabel={T},
        grid=both, 
        axis lines=middle, 
        xtick=\empty, ytick=\empty,
        ymin=0, ymax=5,
        xmin=0, xmax=6,
        legend style={at={(0.5,-0.2)},anchor=north}
    ]
    % Line plot: positive ramp, horizontal, positive ramp, horizontal, negative ramp
    \addplot[thick, darkblue] coordinates {
        (0,0.1) 
        (1,2)  % Positive ramp
        (2,2)       % Horizontal line
        (3,4)       % Positive ramp
        (4,4)       % Horizontal line
        (5,0.2)      % Negative ramp
    };
    \node at (axis cs: 3, 1.4) {Cure Cycle};
    \end{axis}
\end{tikzpicture}
};

% --- Nodes ---
% Inputs to the PINO Model
\node (ht) at(-6.25,-2) [process1] {$h_{top}$};
\node (hb) at(-5,-2) [process1] {$h_{bottom}$};
\node (Lt) at(-3.75,-2) [process1] {$L_{t}$};
\node (plus) at(-5,-1.4) {$+$};

% Curly bracket around the three nodes
\draw [decorate,decoration={brace,amplitude=8pt,mirror}] 
  (ht.south west) -- (Lt.south east) node [black,midway,xshift=0.4cm]{};
\node (designtitle) at(-5,-3.3) {\textit{Tool and Equipment}};
\node (designtitle) at(-5,-3.6) {\textit{Design Variable}};
\draw[->, thick] (-3,-1.25) -- (-1,-1.25);

% PINO Model box
\fill[white, opacity=0.7 , rounded corners=10pt] (-1,-4) rectangle (3.5,1.5);
% Sub-rectangle 1 (with only the border in maroon)
\draw [rounded corners=5pt, thick, draw=plightred] (-0.5,-1.1) rectangle (1.5,1.4);

%\draw[->, thick] (-0.8,0.15) -- (-0.5,0.15);  % Arrow  to Sub-rectangle

% Sub-rectangle 2 (with only the border in maroon)
\draw[rounded corners=5pt, thick, draw=pdarkred] (-0.5,-3.9) rectangle (1.5,-1.4);

%\draw [->, thick] (-0.8,-2.65) -- (-0.5,-2.65);  % Arrow  to Sub-rectangle

% Sub-rectangle 3 (with only the border in maroon)
\draw[rounded corners=3pt, thick, draw=pblue] (2.2,-2.3) rectangle (2.9,0.1);

\draw[->, thick] (1.5, 0.15) -- (1.8, 0.15) -- (1.8, -1.25) -- (2.2, -1.25);

\draw[->, thick] (1.5, -2.65) -- (1.8, -2.65) -- (1.8, -1.25) -- (2.2, -1.25);

%\draw [->, thick] (2.9,-1.25) -- (3.3,-1.25);  % Arrow  to Sub-rectangle

% Circles in Sub-rectangle 1 (3 in the first row)
\node (Bc1) [draw, circle, minimum size=0.2cm] at (0,0.8) {$\sigma$};
\node (Bc2) [draw, circle, minimum size=0.2cm] at (1,0.8) {$\sigma$};

% Circles in Sub-rectangle 2 (3 in the second row)
\node (Bc3)[draw, circle, minimum size=0.2cm] at (0,-0.5) {$\sigma$};
\node (Bc4) [draw, circle, minimum size=0.2cm] at (1,-0.5) {$\sigma$};

% Circles in Sub-rectangle 3 (2 in the column )
\node (c1)[draw, circle, minimum size=0.2cm] at (2.55,-0.7) {$\sigma$};
\node (c2) [draw, circle, minimum size=0.2cm] at (2.55,-1.7) {$\sigma$};

% Arrows from each circle in the first row to the other circles
\draw[->, thick] (Bc1) -- (Bc2); % Arrow from Tc1 to Tc2
\draw[->, thick] (Bc1) -- (Bc4); % Arrow from Tc1 to Tc4
\draw[->, thick] (Bc3) -- (Bc2); % Arrow from Tc3 to Tc2
\draw[->, thick] (Bc3) -- (Bc4); % Arrow from Tc3 to Tc4

% Circles in Sub-rectangle 1 (3 in the first row)
\node (Tc1) [draw, circle, minimum size=0.2cm] at (0,-2) {$\sigma$};
\node (Tc2) [draw, circle, minimum size=0.2cm] at (1,-2) {$\sigma$};

% Circles in Sub-rectangle 2 (3 in the second row)
\node (Tc3)[draw, circle, minimum size=0.2cm] at (0,-3.3) {$\sigma$} ;
\node (Tc4) [draw, circle, minimum size=0.2cm] at (1,-3.3) {$\sigma$};

% Arrows from each circle in the first row to the other circles
\draw[->, thick] (Tc1) -- (Tc2); % Arrow from Tc1 to Tc2
\draw[->, thick] (Tc1) -- (Tc4); % Arrow from Tc1 to Tc4
\draw[->, thick] (Tc3) -- (Tc2); % Arrow from Tc3 to Tc2
\draw[->, thick] (Tc3) -- (Tc4); % Arrow from Tc3 to Tc4

% Optimizer
\fill[white, opacity=0.7,rounded corners=10pt] (5,-4) rectangle (7,1.5);
\draw[->, thick] (3.5,-1.25) -- (5,-1.25);
\node[rotate=0] at (6,-0.8) {\textbf{Adam}};
\node[rotate=0] at (6,-1.5) {\scalebox{3}{\textbf{\faCog}}};

\draw[->, thick] (7,-1.25) -- (9,-1.25);
% Optimize Design
\fill [white, opacity=0.7, rounded corners=10pt] (9,-4) rectangle (13,1.5);
\node (plot) at (11,0) {
\begin{tikzpicture}
    \begin{axis}[
        width=5cm, height=4cm,
        xlabel={t}, ylabel={T},
        grid=both, 
        axis lines=middle, 
        xtick=\empty, ytick=\empty,
        ymin=0, ymax=5,
        xmin=0, xmax=6,
        legend style={at={(16.5,15.8)},anchor=north}
    ]
    % Line plot: positive ramp, horizontal, positive ramp, horizontal, negative ramp
    \addplot[thick, maroon] coordinates {
        (0,0.1) 
        (1,3)  % Positive ramp
        (2,3)       % Horizontal line
        (3,4.5)       % Positive ramp
        (4,4.5)       % Horizontal line
        (5,0.2)      % Negative ramp
    };
    \node at (axis cs: 3, 1.4) {Cure Cycle};
    \end{axis}
\end{tikzpicture}
};

% --- Nodes ---
% Inputs to the PINO Model
\node (ht) at(9.75,-2) [process2] {$h_{top}$};
\node (hb) at(11,-2) [process2] {$h_{bottom}$};
\node (Lt) at(12.25,-2) [process2] {$L_{t}$};
\node (plus) at(11,-1.4) {$+$};

% Curly bracket around the three nodes
\draw [decorate,decoration={brace,amplitude=8pt,mirror}] 
  (ht.south west) -- (Lt.south east) node [black,midway,xshift=0.4cm]{};
\node (designtitle) at(11,-3.3) {\textit{Tool and Equipment}};
\node (designtitle) at(11,-3.6) {\textit{Design Variable}};

%

% Add feedback line
\draw[->, thick] (6, -4) -- (6, -4.5) -- (-2, -4.5) -- (-2, -1.25);
\node (feedbacktext) at(2,-4.3) {\textit{Updated design}};
\node (loss1) at(4.2,-1) {\textit{Evaluate}};
\node (loss2) at(4.2,-1.5) {\textit{losses}};

% captions
\node (Initial Design) at(-5,1.7) {\textit{Initial Design}};
\node (PIDON) at(1,1.7) {\textit{PIDON Model}};
\node (Optimizer ) at(6,1.7) {\textit{Optimizer}};
\node (Optimum Design ) at(11,1.7) {\textit{Optimal Design}};

\end{tikzpicture}
\caption{An accelerated gradient-based design optimization framework: Using the PIDON model, the framework predicts the spatio-temporal evolution of part temperature and DOC, calculates the loss, and iteratively adjusts design variables through gradient-based optimization to find the optimal design.}
\label{fig:pino_framework}
\end{figure}

We developed an AI-driven accelerated design optimization framework for composite synthesis in autoclaves, as illustrated in Figure \ref{fig:pino_framework}. This framework integrates a generalized, accurate, and computationally efficient PIDON model, serving as a differentiable simulator to optimize the curing process. By leveraging the zero-shot super-resolution capabilities of PIDON across diverse design parameters and its differentiability for gradient-based optimizers like Adam, our framework enables rapid and efficient exploration of design variables.
\subsubsection{Design Tasks}
Several studies have focused on optimizing the cure cycle for composite synthesis to achieve various objectives. For instance, one study aimed to minimize residual stress and ensure uniform curing \cite{shah2018optimal},while another focused on maximizing the DOC, controlling peak temperature, minimizing post-gelation gradients, and reducing curing time \cite{vafayan2015development}. 

This study aims to optimize the cure cycle of AS4/8552 composites by balancing mechanical performance and structural integrity through four key objectives. The first objective is achieving a desired DOC within the range $0.85 < \alpha \big|_{t=t} < 0.95$, ensuring optimal mechanical properties while avoiding brittleness \citep{rothenhausler2023interplay}. Second, minimizing DoC gradients $\frac{\partial \alpha}{\partial x} \big|_{t=t}$ is critical for reducing residual stress and shrinkage \citep{yuan2021multi}. Third, maximum part temperature (exotherm) must remain below 185 $^{\circ}\text{C}$ to prevent thermal degradation \citep{fabris2018framework}. Finally, controlling thermal lag ($\Delta T < 20\,^{\circ}\text{C}$) ensures even curing by limiting temperature differences between autoclave air and composite parts \citep{fabris2018framework}. These objectives are essential to optimize curing for performance and material integrity.
\subsubsection{Problem Formulation}
As elaborated in subsection \ref{composites}, we consider nine design variables (i.e., six parameters related to the cure cycle and three associated with tool/equipment design) represented as: $u = [r_1, r_2, hd_1, hd_2, ht_1, ht_2, h_{top}, h_{bot}, L_t]$. The optimization problem is formulated to minimize a total loss function \(\mathcal{L}(u)\), composed of four individual loss terms representing the objectives defined in the above subsection:
\begin{equation}
  \mathcal{L}(u) = \mathcal{L}_{1}(u) +  \mathcal{L}_{2}(u) + \mathcal{L}_{3}(u) +\mathcal{L}_{4}(u)  
\end{equation}
Using the PIDON model as a surrogate, we predict part temperature and DOC, used in the loss calculations. For the first objective, a penalty function is defined based on the desired DOC. A penalty function ensures the DOC remains within the desired range of 0.85 to 0.95. If the DOC falls outside this range: either the minimum of DOC across the laminate is below 0.85 or the maximum of DOC across the laminate exceeds 0.95, a penalty is applied; otherwise, the loss is zero.This can be expressed as:
\begin{equation}
\mathcal{L}_1(u) = 
\begin{cases} 
\left| 0.85 - \min(G^{\alpha}_{\theta}(u)(y_{t})) \right|, & \text{if } \min(G^{\alpha}_{\theta}(u)(y_{t})) < 0.85; \\
\left| \max(G^{\alpha}_{\theta}(u)(y_{t})) - 0.95 \right|, & \text{if } \max(G^{\alpha}_{\theta}(u)(y_{t})) > 0.95; \\
0, & \text{else}.
\end{cases}
\end{equation}
Here, \( y_t = [t|_{t=t}, z] \) and \( z \) is the collocation points in the spatial domain \( z \). A second objective is to minimize the gradient of the DOC across the laminate at the end of the curing process. Since the PIDON model is differentiable, the gradient is computed using automatic differentiation across the laminate thickness, and it is then averaged. This can be expressed as follows:
\begin{equation}
\mathcal{L}_2(u) = \frac{1}{N_z} \sum_{i=1}^{N_z} \left| \frac{\partial (G^{\alpha}_{\theta}(u)(y_{t}^{(i)}))}{\partial z}  \right|
\end{equation}
Here, \( N_z \) is the number of collocation points along the \( z \)-coordinate. The third objective ensures that the maximum part temperature does not exceed 185°C. Hence, a penalty function is introduced as follows:
\begin{equation}
\mathcal{L}_3(u) = \begin{cases} 
\left| 185 - \max(G^{T_{t}}_{\theta}(u)(y)) \right|, & \text{if } \max(G^{T_{t}}_{\theta}(u)(y)) > 185; \\
0, & \text{else.}
\end{cases}
\end{equation}
The fourth objective aims to limit the thermal lag to values below 20°C. This objective can be expressed as follows:
\begin{equation}
\mathcal{L}_4(\mathbf{\phi}) = \frac{1}{N} \sum_{i=1}^{N} \begin{cases} 
\left| 20 - (T_{a}(y^{(i)}) - G^{T_{t}}_{\theta}(\mathbf{\phi})(y^{(i)})) \right|, \\
\text{if } (T_{a}(y^{(i)}) - G^{T_{t}}_{\theta}(\mathbf{\phi})(y^{(i)})) > 20; \\
0, \text{else.}
\end{cases}
\end{equation}
\subsubsection{Optimizer}
The optimization problem for the inverse design of composites autoclave processing is formulated as:
as follows:
\begin{equation}
\begin{aligned}
    u^* &= \arg \min_{u} \mathcal{L}(u), \\
    \text{subject to:} \quad &u_{\text{min}} \leq u \leq u_{\text{max}}.
\end{aligned}
\label{eq:optimization_problem}
\end{equation}
The bounds \(u_{\text{min}}\) and \(u_{\text{max}}\) define the feasible design space for the design variables \(u\) and are summarized in Table \ref{tab:designRange}. They are typically defined by the inherent physical constraints of the process, such as the range of HTCs and temperature achievable by the autoclave oven. These constraints also align with the ranges of the input functions used during the training of the PIDON model. By adhering to these bounds, the design variables remain within the range of values on which the PIDON model was trained, ensuring it provides accurate and reliable predictions. The solution \(u^*\) represents the optimal set of design parameters that minimizes the total loss \(L(u)\),  while satisfying all objectives, constraints, and bounds. 

To solve this optimization problem, we employ a gradient-based approach, leveraging the differentiability of the PIDON model. Gradient-based methods are particularly advantageous for differentiable objective functions, as they use gradient information to efficiently explore the design space \citep{allen2022physical,um2020solver}. In this case, the loss function depends on the output of PIDON, which itself is a function of the input (i.e., design variables). By leveraging automatic differentiation, we can compute the gradient of the loss with respect to the design variables by propagating the gradients through the PIDON using the chain rule. These methods converge with fewer function evaluations, making them computationally efficient compared to sampling-based or heuristic approaches \citep{allen2022physical}. Specifically, we apply the Adam optimizer \citep{kingma2014adam}, which adapts the learning rate during optimization, improving the convergence speed and stability.

The optimization process involves the following steps (Figure \ref{fig:pino_framework}):
\begin{enumerate}
    \item \textbf{Initial Guess:} Select the initial design parameters $u^{0}$, either randomly or based on domain knowledge, and pass them to the PIDON model.
    \item \textbf{Forward Prediction:} Use the PIDON model to predict the spatio-temporal evolution of part temperature and DOC across the laminate based on the current design variables.
    \item \textbf{Loss Evaluation:} Compute the individual loss functions based on the predicted temperature and DOC values.
    \item \textbf{Gradient Computation:} Utilize automatic differentiation to compute the gradients of the total loss function \(\mathcal{L}(u)\) with respect to the design parameters \(u\). This step involves back-propagating the gradients through the PIDON model.
    \item \textbf{Update Design Variables:} Apply a gradient-based optimizer (e.g., Adam) to update the design variables \(u\) in the direction that minimizes the loss function using the computed gradients.
\end{enumerate}
Design variables are clipped within the specified bounds presented in Table \ref{tab:designRange} to prevent erroneous predictions. If a variable exceeds the upper limit, it is clipped to the upper value; if it falls below the lower bound, it is clipped to the lower limit. Each loss function is normalized to maintain consistent scales and prevent dominance of any objective. To compare efficiency, we implemented Nesterov-Adam (NAdam), PSO, and GA, using the same PIDON model and loss calculation procedure as the Adam optimizer for consistency.
\section{Results and Discussion}
\subsection{Validation of PIDON Model}
\begin{figure}[]  
    \centering
    \includegraphics[width=0.48\textwidth]{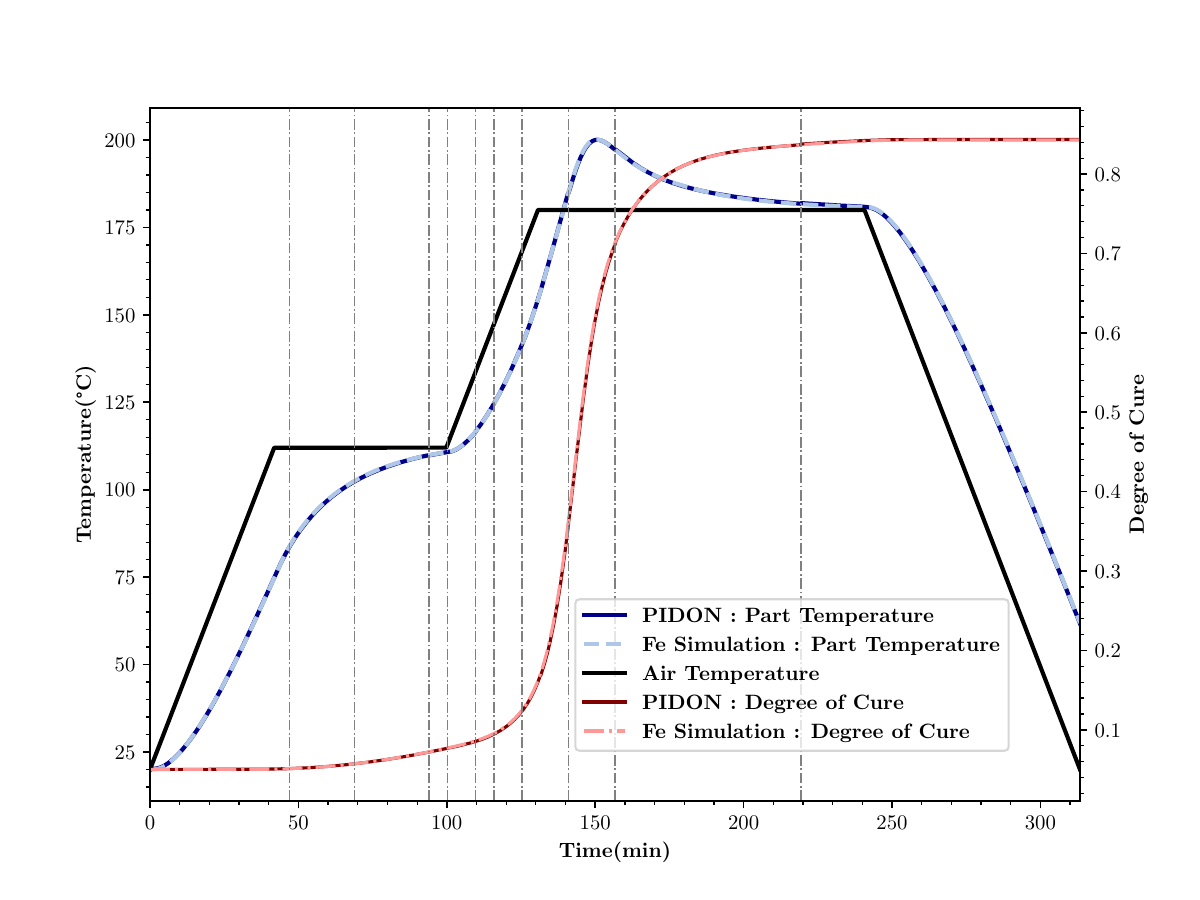} 
    \caption{Comparison of predicted part temperature and DOC from the PIDON model with FE simulation at the midpoint of the composite.}     % Caption for the figure
    \label{fig:pino_pred}                             % Label for referencing the figure
\end{figure}

To train the PIDON model, the time domain was divided into 11 subdomains with various lengths, accommodating the level of complexity within each subdomain. All PIDON models used an identical architecture, as detailed in Table \ref{tab:hyperparameter}. We trained the model with 600 random input parameter combinations (i.e., input functions) spanning a wide range of values (Table \ref{tab:designRange}). We evaluated the model against finite element simulations as ground truth across five different test cases. Specifically, the part temperature and degree of cure (DOC) at the midpoint of the composite for a given set of input functions are compared with FE simulation results, as illustrated in Figure \ref{fig:pino_pred}. The plot demonstrates that the PIDON model provides highly accurate predictions, with its results closely overlapping those of the FE simulations. We also benchmarked its performance against previously developed operator-based models for composites processing, namely the Physics-Informed Neural Operator (PINO) \cite{ramezankhani2025advanced} and the physics-informed Fourier Neural Operator (FNO) \cite{meng2023novel} (Table \ref{tab:pidon_comparison}). We used mean absolute error (MAE) and mean maximum error as the evaluation metrics to compare model predictions with finite element simulations. The MAE was computed by averaging the absolute errors within and across all test cases. The mean maximum error was obtained by averaging the maximum errors from each test case. The results clearly show that the PIDON outperforms both PINO and FNO models, achieving MAE and mean maximum error that are approximately 50\% lower than those reported for the other models, despite being trained on a wider range of input functions. This demonstrates the superior accuracy and reliability of the proposed PIDON approach for predicting temperature and DOC in composite systems.

\begin{table}[]
\centering
\caption{Comparison of our proposed PIDON (ours), PINO \cite{ramezankhani2025advanced}, and FNO \cite{meng2023novel} based on mean absolute error(MAE) and maximum error (MAX) for temperature and DOC predictions.}
\begin{tabular}{@{}c|cc|cc|cc@{}}
\toprule
\multirow{2}{*}{Variable} & \multicolumn{2}{c|}{PIDON (ours)}     & \multicolumn{2}{c|}{PINO}      & \multicolumn{2}{c}{FNO}        \\ \cmidrule(l){2-7} 
                          & \multicolumn{1}{c|}{MAE} & MAX & \multicolumn{1}{c|}{MAE} & MAX & \multicolumn{1}{c|}{MAE} & MAX \\ \midrule
$T_{c}$(°C) & \multicolumn{1}{c|}{0.189} & 0.94  & \multicolumn{1}{c|}{0.362} & 2.231 & \multicolumn{1}{c|}{0.226} & 3.2  \\ \midrule
$\alpha$    & \multicolumn{1}{c|}{0.001} & 0.008 & \multicolumn{1}{c|}{0.002} & 0.02  & \multicolumn{1}{c|}{0.008} & 0.03 \\ \bottomrule
\end{tabular}
\label{tab:pidon_comparison}
\end{table}

\begin{table}[]
\centering
\caption{Performance comparison between gradient-based (i.e., Adam and NAdam)  and gradient-free  (i.e., PSO and GA) optimization models, highlighting computational time, PIDON Model calls and optimization objective metrics. Gradient-based optimizer results are averaged over 10 initial guesses, while gradient-free results use a single run. Model calls indicate the total number \texttt{(forward + backward)} function calls per optimization iteration.}
\begin{tabular}{@{}c|c|c|c|c|c|c@{}}
\toprule
Optimizer &
  \begin{tabular}[c]{@{}c@{}}Time \\ (Min.)\end{tabular} &
  \begin{tabular}[c]{@{}c@{}}Model Calls \\ (per iteration)\end{tabular} &
  \begin{tabular}[c]{@{}c@{}}Mean DOC \\ Gradient\end{tabular} &
  \begin{tabular}[c]{@{}c@{}}Max. $T_c$\\(=\textless{}185)\end{tabular} &
  \begin{tabular}[c]{@{}c@{}}Mean \\ Thermal Lag\\ (=\textless{}20)\end{tabular} &
  \begin{tabular}[c]{@{}c@{}} Mean DOC\\ (\textgreater{}=0.85)\end{tabular} \\ \midrule
\textit{Adam}  & 20.37 & \(1+2\)  & 0.0044±0.0002 & 185.21±0.25 & 14.10±0.67 &  0.853±0.0002      \\ \midrule
\textit{NAdam} & 20.02 & \(1+2\) &  0.0043±0.0001 & 185.12±0.16 & 14.16±0.61 &  0.852±0.00008       \\ \midrule
\textit{PSO}   & 58.7  & \(10 \times (1+1)\) & 0.0057        & 185.18      & 14.11      & 0.847  \\ \midrule
\textit{GA}    & 69.03 & \(100 \times (1+1)\) & 0.0045        & 185.07      & 14.91 & 0.852  \\ \bottomrule
\end{tabular}
\label{tab:optimizer_comparison}
\end{table}
\subsection{Design Optimization via Differentiable Neural Operator}
\begin{figure*}[t]  
    \centering
    \includegraphics[width=1\textwidth]{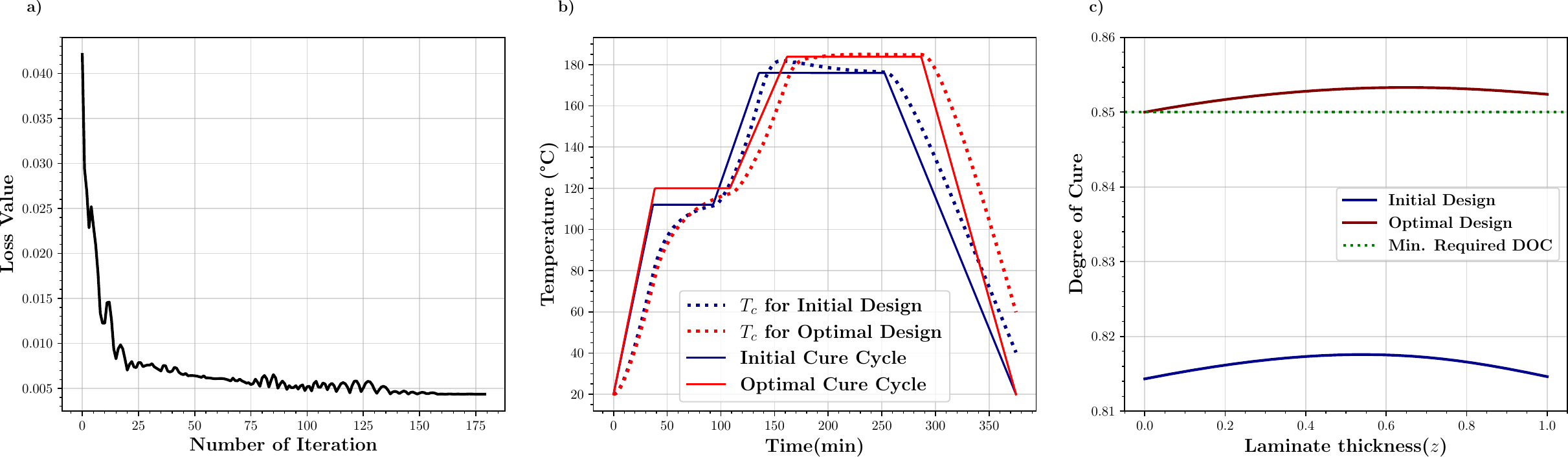} 
    \caption{a) Evolution of the total loss during each iteration of the optimization process using the Adam optimizer. b) Comparison of the initial and optimized cure cycle profiles using the Adam optimizer, including the part temperature ($T_c$) behavior at the midpoint. c) DOC evolution across the laminate thickness from the initial (blue) to optimized (red) design variables. All results correspond to the optimization process with an initial guess for the design parameters \(u^{0} = [2.5, 1.5, 56, 117, 112, 176, 75, 100, 2.2]\).}
    \label{fig:loss_cure_cyce}                             % Label for referencing the figure
\end{figure*}
\begin{figure*}[t]  
    \centering
    \includegraphics[width=1\textwidth]{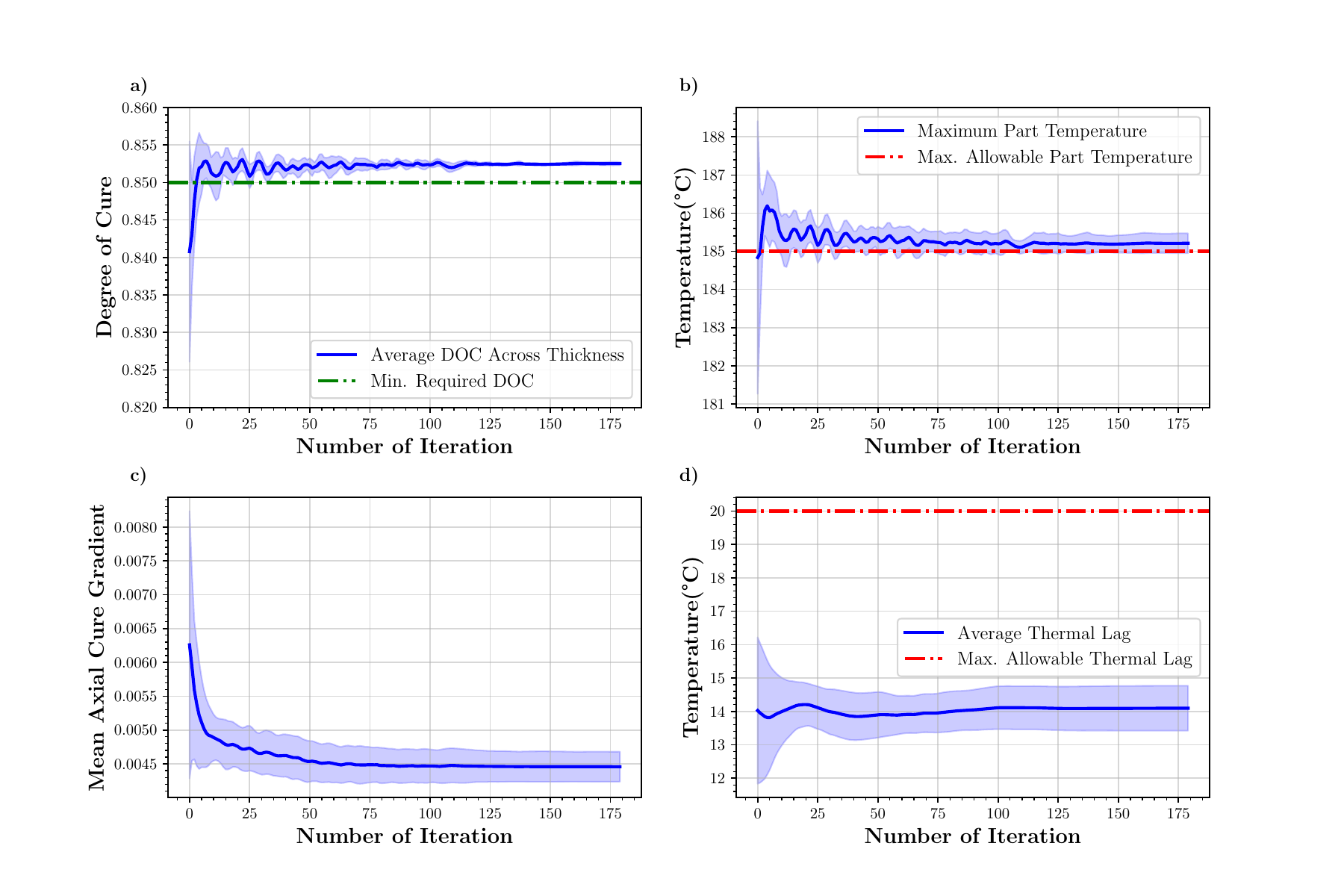} 
    \caption{Evolution of key metrics during optimization: a) Average DOC across the laminate thickness, b) Maximum Temperature observed in the laminate, c) Average Axial Cure Gradient, and d) Average Thermal Lag, at each optimization step. Results are averaged over 10 different initial guesses, with shaded bands representing ± standard deviation to indicate variability across the optimization runs.}     % Caption for the figure
    \label{fig:constraints}                             % Label for referencing the figure
\end{figure*}
A trained PIDON model is used as a surrogate model to identify the optimal design parameters for the autoclave curing process. Losses are evaluated using a 20×100 grid of collocation points, with 20 points in the spatial domain ($z$) and 100 in the time domain. The performance of the framework was evaluated for two composite part thicknesses: 20 mm, as discussed in this section, and 30 mm, which is detailed in Appendix \ref{30mm}. Figure \ref{fig:loss_cure_cyce} shows the key outcomes of the optimization procedure for a random initial state \(u^{0} = [2.5, 1.5, 56, 117, 112, 176, 75, 100, 2.2]\). Figure \ref{fig:loss_cure_cyce}.a shows the evolution of the total loss during each iteration of the optimization process using the Adam optimizer. Figure \ref{fig:loss_cure_cyce}.b compares the initial and optimized cure cycle profiles, including the part temperature behavior at the midpoint of the composite. Although the air temperature is higher for the optimized design compared to the initial design, the maximum part temperature remains below the threshold in both cases. The initial and optimal DOC distributions across the laminate thickness are compared in Figure \ref{fig:loss_cure_cyce}.c. In the initial design, the DOC remains way below the 0.85 threshold, while the optimized design achieves a DOC exceeding 0.85 across the entire laminate thickness. These results clearly demonstrate the effectiveness of the proposed framework in achieving substantial improvements in cure cycle performance. The optimization framework yields the optimized design variables $u^{*} = [2.61, 1.2, 70, 125, 119, 183.9, 52.68, 91.61, 2.75]$ which satisfies all objectives, ensuring a high-quality manufactured part. The total computation time for a single optimization run is 20.37 minutes.

The obtained optimal design aligns very well with the theoretical expectations. Specifically, a higher value of \(r_1\) and a lower value of \(r_2\) ensures efficient heating during the first ramp phase while preventing excessive heat buildup (caused by resin polymerization) during the second phase. The optimal design has the highest \(ht_1\) to maximize heat during the first hold, ensuring a cure rate above 0.85 while avoiding excess heat in the second hold. For \(L_t\), the optimal design has a mid-range value. While a higher tool thickness can help mitigate excessive exothermic reactions, it may lead to a non-uniform DOC across the laminate, thus the chosen thickness can balance these effects.

% In addition, the total computation time for a single optimization run is 20.37 minutes, highlighting the framework's efficiency.

Further, to demonstrate the robustness of our framework, we present the optimization results using Adam for 10 different initial guesses. The average objective evolution across these runs is shown in Figure \ref{fig:constraints}. The figure illustrates the progression of key objectives during optimization, including the average DOC across the laminate (Figure \ref{fig:constraints}.a), maximum part temperature (Figure \ref{fig:constraints}.b), average axial cure gradient (Figure \ref{fig:constraints}.c), and average thermal lag (Figure \ref{fig:constraints}.d). The blue line represents the mean value across the 10 initial guesses, while the shaded band indicates the standard deviation. From the figure, it is evident that in some initial designs, the DOC is below 0.85. However, the optimal designs consistently achieve a DOC exceeding 0.85 across the laminate thickness. Additionally, the variation in DOC across the thickness is significantly reduced in the optimized designs compared to the initial designs. This improvement is further reflected in the steady decrease of the DOC gradient with each optimization step. Since DOC and maximum part temperature are correlated, initial designs with higher DOC exhibit higher part temperatures. In the optimization process, the maximum part temperature, which initially varies widely, stabilizes precisely at 185°C in the optimized designs. Lastly, the average thermal lag successfully remains below 20°C throughout the optimization procedure.
\subsection{Comparison to gradient-free methods}
In addition to our proposed optimization framework, we implemented and compared alternative optimization techniques, including NAdam, PSO, and GA, to evaluate their performance. The NAdam was executed for 180 iterations. The PSO was implemented with a swarm of 20 particles, running for 25 iterations, while the GA utilized a population size of 100 and was executed for 100 iterations. The performance of these optimizers, along with their computation times, is summarized in Table \ref{tab:optimizer_comparison}. For gradient-based optimizers, such as Adam and NAdam, the results are averaged over 10 different initial guesses to assess robustness. For population-based optimizers, since they are less sensitive to initial guesses, a single optimization run was executed and reported. In terms of the achieved optimal states, the gradient-based and gradient-free methods resulted in comparable performance. In terms of the number of function calls during the optimization process, gradient-based methods call the PIDON model once for forward prediction and twice for backward propagation (axial DOC gradient calculation and optimizer gradient update) per iteration (1 + 2). In contrast, gradient-free methods require one forward and one backward call for axial DOC gradient calculation for each individual in the population per iteration \((1+1) \times p\), where \(p\) is the population size. This makes gradient-based optimizers significantly more efficient with fewer function evaluations during the optimization process. Hence, the gradient-based optimizers demonstrated a significant computational advantage, converging approximately \textit{three times} faster than their gradient-free counterparts. This clearly indicates the advantage of utilizing a differentiable neural operator model in conjunction with a gradient-based optimization approach for accelerated and robust design optimization in advanced manufacturing and materials processing applications.
\section{Conclusions}
In this work, we present an end-to-end accelerated design optimization framework for advanced engineering design, specifically developed for the autoclave curing process in composite materials processing. At the core of the framework is an enhanced PIDON model, specifically designed to capture the nonlinear dynamics of the autoclave process with high accuracy. The improved PIDON integrates a nonlinear decoder and temporal domain decomposition techniques, delivering superior performance compared to existing neural operator-based models across wide range of diverse input functions and extended temporal domains. By leveraging the differentiability of the trained PIDON model, the framework utilizes the gradient-based Adam optimizer to efficiently identify optimal design parameters. When benchmarked against gradient-free methods such as PSO and GA, the Adam optimizer achieves comparable performance while operating approximately three times faster. This enables the accelerated and accurate identification of optimal design states for the autoclave curing process. Furthermore, the framework is highly adaptable and can be extended to address other advanced engineering design challenges, making it well-suited for digital twin applications.

%\bibliographystyle{plain}
%\bibliography{references}  %%% Uncomment this line and comment out the ``thebibliography'' section below to use the external .bib file (using bibtex) .





% \end{thebibliography}
\appendix
\section{Hyper-parameters of PIDON and Optimizers}
The hyperparameters for each sub-PIDON model are summarized in Table \ref{tab:hyperparameter}, including network architectures, optimizer settings, training parameters, and hardware specifications. Additionally, The range of input functions for which PIDON is trained is presented in Table \ref{tab:designRange}.

For the Adam optimizer, a custom dynamic learning rate mechanism is implemented. If the loss does not improve for a specified number of iterations (defined by \texttt{patience}), the learning rate is reduced by the factor \texttt{reduction\_factor}, with a lower bound set by \texttt{min\_lr}. The optimizer starts with an initial learning rate of \(0.01\), performs up to 180 steps, and reduces the learning rate by a factor of 0.5 if no improvement is observed after 10 iterations, with the learning rate not dropping below \(1 \times 10^{-5}\). A similar dynamic learning rate mechanism is employed for the Nadam optimizer. The GA used in this study was configured with a population size of 10, a mutation probability of 0.1, an elitism ratio of 0.01, and a crossover probability of 0.5. The selection process retained the top 30\% of individuals as parents, applying uniform crossover for offspring generation. The PSO algorithm was set with a cognitive parameter of 0.5, a social parameter of 0.3, and an inertia weight of 0.9.
\begin{table}[htbp]
\centering
\begin{minipage}{0.45\textwidth}
\centering
\caption{Hyperparameters of the PIDON model.}
\begin{tabular}{@{}c|c@{}}
\toprule
Branch Network        & [50, 50, 50]        \\ \midrule
Trunk Network         & [50, 50, 50, 50, 50] \\ \midrule
Non-linear Decoder    & [50, 50, 50, 50]     \\ \midrule
Optimizer             & Adam                             \\ \midrule
Initial Learning Rate & \(1 \times 10^{-3}\)             \\ \midrule
Learning Rate Decay & \begin{tabular}[c]{@{}c@{}}Decay rate of 0.9 \\ every 1000 steps\end{tabular}  \\ \midrule
Training Epochs       & 200                              \\ \midrule
Training Library      & JAX                              \\ \midrule
Hardware            & \begin{tabular}[c]{@{}c@{}}NVIDIA T4 GPU with \\ 104 GB of memory\end{tabular} \\ \bottomrule
\end{tabular}
\label{tab:hyperparameter}
\end{minipage}
\hfill
\begin{minipage}{0.45\textwidth}
\centering
\caption{Design variables (input functions) and their corresponding ranges.}
\begin{tabular}{@{}c|c@{}}
\toprule
Design Parameter                                     & Range          \\ \midrule
$r_1$ (°C/min)                                       & {[}1.2, 3{]}   \\ \midrule
$r_2$ (°C/min)                                       & {[}1.2, 3{]}   \\ \midrule
$hd_1$ (min)                                         & {[}50, 70{]}   \\ \midrule
$hd_2$ (min)                                         & {[}115, 125{]} \\ \midrule
$ht_1$ (°C)                                          & {[}100, 120{]} \\ \midrule
$ht_2$ (°C)                                          & {[}175, 185{]} \\ \midrule
$h_{top}$ ($W/m^{2}K$)                               & {[}70, 120{]}  \\ \midrule
$h_{bot}$ ($W/m^{2}K$)                               & {[}40, 90{]}   \\ \midrule
$L_t$ (cm)                                           & {[}2, 4{]}     \\ \bottomrule
\end{tabular}
\label{tab:designRange}
\end{minipage}
\end{table}
\section{Additional Results}
\label{30mm}
\begin{figure}[t]  
    \centering
    \includegraphics[width=0.8\textwidth]{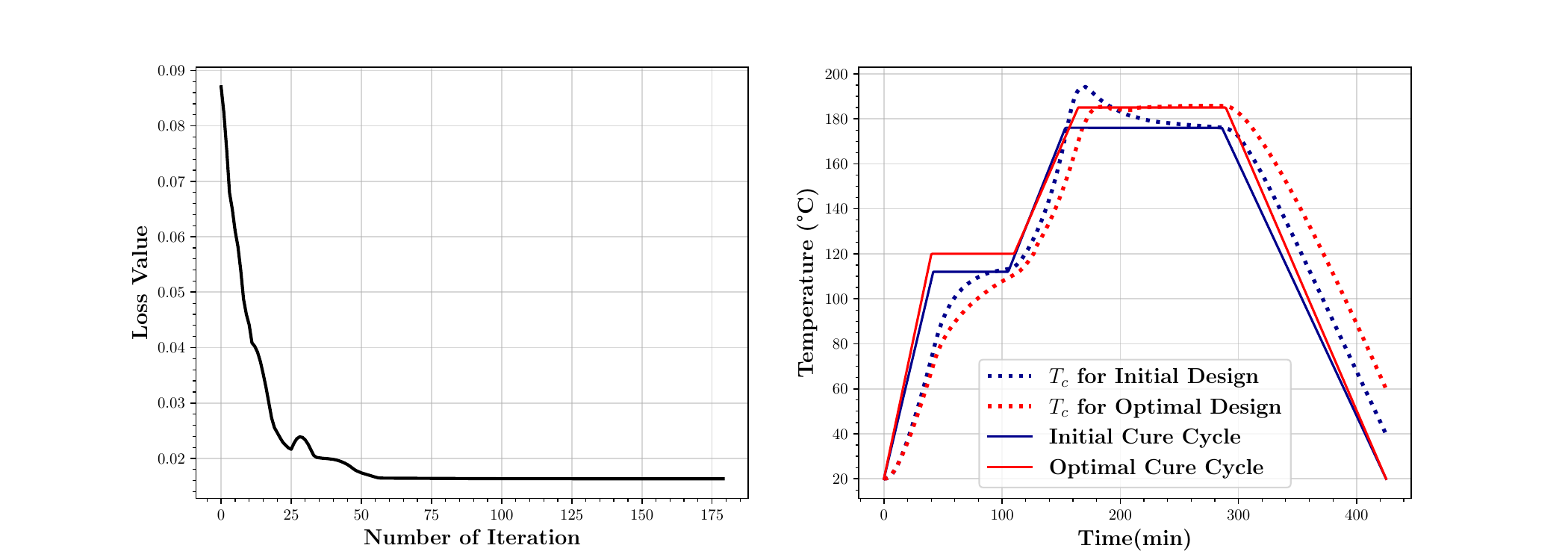} 
    \caption{Optimization results for 30 mm thick composite part: a)  Evolution of the total loss during each iteration of the optimization process using the Adam optimizer b) Comparison of the initial and optimized cure cycle profiles using the Adam optimizer, and the corresponding part temperature behavior at the midpoint.}     % Caption for the figure
    \label{fig:30mm_loss_comp}                             % Label for referencing the figure
\end{figure}

\begin{figure}[t]  
    \centering
    \includegraphics[width=0.8\textwidth]{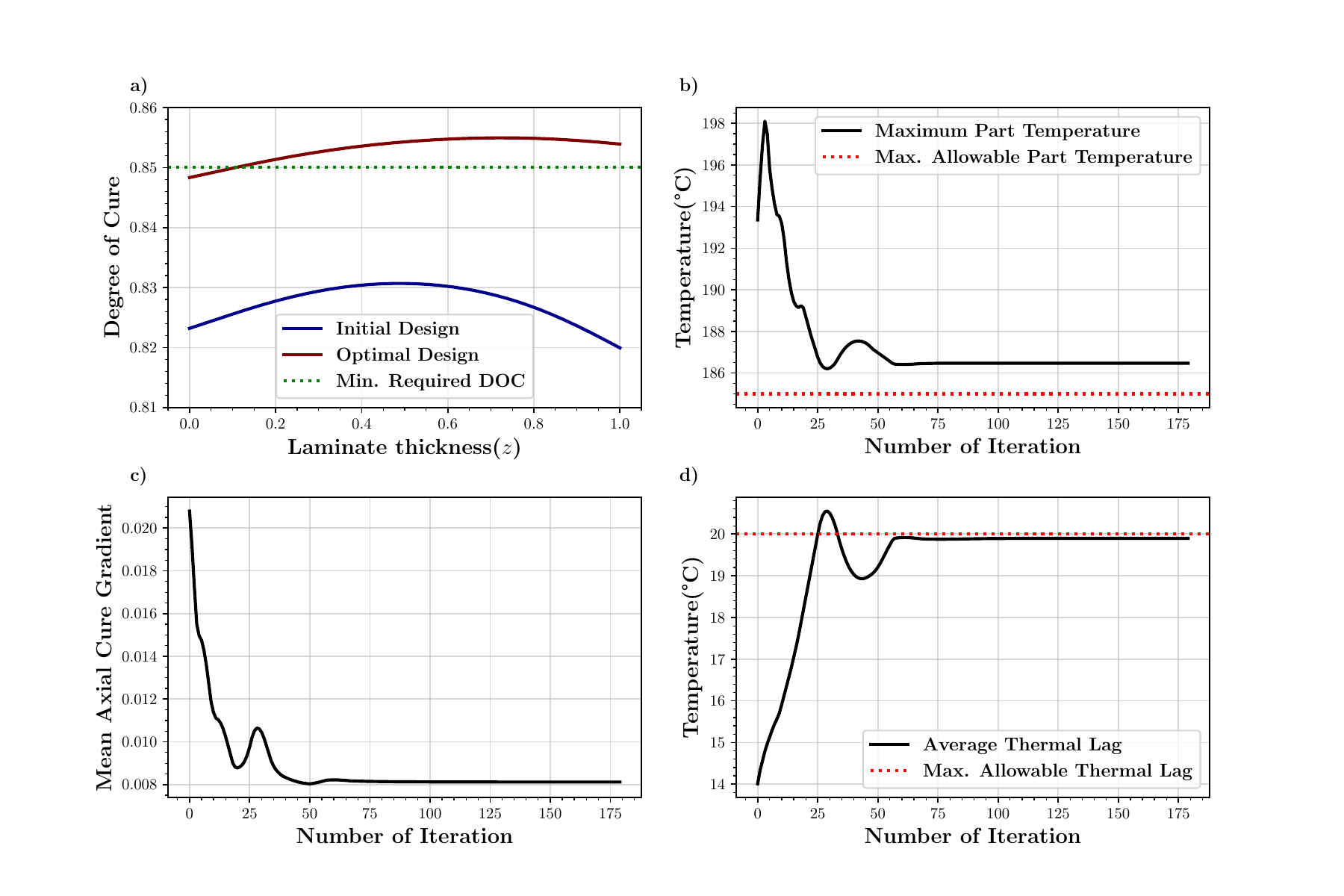} 
    \caption{For 30mm thick material: a)  DOC evolution across the laminate thickness from the initial (blue) to optimized (red) design variables. Evolution of b) maximum temperature (exotherm). c) average axial (through-thickness) cure gradient, and d) average thermal lag evolution at each optimization step.}     % Caption for the figure
    \label{fig:constraints_exotherm}                             % Label for referencing the figure
\end{figure}

To demonstrate the robustness of the proposed framework in addressing challenging scenarios, this section presents results for a 30 mm thick composite part. The optimization problem becomes significantly more challenging for the 30 mm thick material due to the excessive exothermic heat generation and the conflicting objectives of minimizing the maximum part temperature while achieving a sufficiently high DOC. The evolution of the total loss during optimization is shown in Figure \ref{fig:30mm_loss_comp}.a, while Figure \ref{fig:30mm_loss_comp}.b compares the initial and optimized cure cycles, along with the corresponding part temperatures at the composite midpoint. The initial cure cycle design exhibits excessive heat accumulation, resulting in a significantly elevated maximum part temperature. In contrast, the optimized design effectively regulates the maximum part temperature, bringing it to approximately 186.4°C—slightly above the target value of 185°C but still well within the acceptable range for practical applications \citep{fabris2018framework}. The DOC distribution acrros the laminate thickness for both the initial and optimized designs is illustrated in Figure \ref{fig:constraints_exotherm}.a. In the initial design, the DOC remains below 0.85, whereas the optimized design achieves a DOC exceeding 0.85 across all regions, except those near the tool. Furthermore, the optimized design demonstrates a substantial reduction in DOC variation, as indicated by the flatter maroon line compared to the darker blue line, which is also reflected in the decreasing DOC gradient over successive iterations in the Figure \ref{fig:constraints_exotherm}.c. Figure \ref{fig:constraints_exotherm}.b depicts the progression of the maximum part temperature during optimization. Initially extremely high, the temperature progressively decreases and stabilizes at approximately 186.4°C in the final design. Similarly, the average thermal lag evolves as shown in Figure \ref{fig:constraints_exotherm}.d. Although it momentarily increases during intermediate steps, the thermal lag ultimately returns to below 20°C in the final optimized solution, aligning with acceptable design criteria. These findings underscore the framework's capability to handle complex, high-dimensional design problems with conflicting objectives, optimizing even for extreme and thick composite materials to yield feasible solutions that satisfy physical constraints and design objectives.
\subsection{Comparison to gradient-free methods}
A comparative analysis of the gradient-based and gradient-free optimization approaches for the 30 mm thick composite material is presented in Table \ref{tab:optimizer_comparison}. The hyperparameters for each optimizer remain consistent with those used for the 20 mm thick material. Notably, gradient-based optimizers demonstrate a threefold speed advantage over their gradient-free counterparts. The optimization process for the 30 mm composite part poses significant challenges due to the conflicting nature of the objectives, including minimizing maximum part temperature and maximizing DOC. This is evident as gradient-free optimizers, such as GA and PSO, struggle to achieve a balanced solution. GA achieves a higher DOC but results in a maximum part temperature exceeding 185°C. Conversely, PSO maintains a maximum part temperature below 185°C but fails to achieve a DOC above 0.85. In contrast, the gradient-based optimizers, Adam and NAdam, demonstrate an ability to effectively balance these competing objectives. While the maximum part temperature for both optimizers slightly exceeds the target value of 185°C, stabilizing at 186.4°C, this result is still within an acceptable range for practical purposes. Additionally, both optimizers achieve a DOC above 0.85, fulfilling the primary performance objectives. This highlights the capability of gradient-based methods in efficiently navigating complex, high-dimensional design spaces.
\begin{table}[]
\caption{Performance comparison between gradient-based (i.e., Adam and NAdam) and gradient-free (i.e., PSO and GA) optimization methods, for a 30 mm thick composite part.}
\centering
\begin{tabular}{@{}c|c|c|c|c|c@{}}
\toprule
Optimizer &
  \begin{tabular}[c]{@{}c@{}}Computation\\ Time (Min.)\end{tabular} &
  \begin{tabular}[c]{@{}c@{}}Mean DOC\\ (\textgreater{}=0.85)\end{tabular} &
  \begin{tabular}[c]{@{}c@{}}Mean DOC \\ Gradient\end{tabular} &
  \begin{tabular}[c]{@{}c@{}}Max. $T_c$\\ (=\textless{}185)\end{tabular} &
  \begin{tabular}[c]{@{}c@{}}Mean \\ Thermal Lag\\ (=\textless{}20)\end{tabular} \\ \midrule
\textit{Adam}  & 19.7 & 0.853 & 0.0083 & 186.4 & 19.53 \\ \midrule
\textit{NAdam} & 19.25 & 0.853 & 0.0084 & 186.38 & 19.51 \\ \midrule
\textit{PSO}   & 58.68  & 0.837 & 0.0098 & 184.98 & 19.11 \\ \midrule
\textit{GA}    & 69.12 & 0.852 & 0.0091 & 188.87 & 15.02 \\ \bottomrule
\end{tabular}
\label{tab:optimizer_comparison_30mm}
\end{table}

\end{document}